\documentclass[letterpaper, 10 pt, conference]{ieeeconf}  

\IEEEoverridecommandlockouts                              

\usepackage{amsmath} 
\usepackage{amssymb}  
\usepackage{booktabs} 
\usepackage{multirow}
\usepackage{graphicx} 
\usepackage{wasysym}
\usepackage{float}

\usepackage{xcolor}
\usepackage[detect-all]{siunitx}

\usepackage{hyperref}
\hypersetup{hidelinks,colorlinks=true,allcolors=black}
\title{\LARGE \bf
MAL: Motion-Aware Loss with Temporal and Distillation Hints \\ for Self-Supervised Depth Estimation
}

\author{Yue-Jiang Dong$^{1}$\quad Fang-Lue Zhang$^{2}$\quad Song-Hai Zhang$^{1*}$\\
\href{https://github.com/YuejiangDong/MAL}{https://github.com/YuejiangDong/MAL}
\vspace{-0.5cm}
\thanks{*corresponding author}\\
\thanks{$^{1}$Yue-Jiang Dong and Song-Hai Zhang are with the Department of Computer Science and Technology, Tsinghua University, Beijing, China
        {\tt\small \{dongyj21@mails.,  shz@\}tsinghua.edu.cn}}
\thanks{$^{2}$Fang-Lue Zhang is with School of Engineering and Computer Science, Victoria University of Wellington, New Zealand
        {\tt\small fanglue.zhang@vuw.ac.nz}
}
}

\begin{document}

\maketitle
\thispagestyle{empty}
\pagestyle{empty}

\begin{abstract}
Depth perception is crucial for a wide range of robotic applications. Multi-frame self-supervised depth estimation methods have gained research interest due to their ability to leverage large-scale, unlabeled real-world data. However, the self-supervised methods often rely on the assumption of a static scene and their performance tends to degrade in dynamic environments. To address this issue, we present Motion-Aware Loss, which leverages the temporal relation among consecutive input frames and a novel distillation scheme between the teacher and student networks in the multi-frame self-supervised depth estimation methods. Specifically, we associate the spatial locations of moving objects with the temporal order of input frames to eliminate errors induced by object motion. Meanwhile, we enhance the original distillation scheme in multi-frame methods to better exploit the knowledge from a teacher network. MAL is a novel, plug-and-play module designed for seamless integration into multi-frame self-supervised monocular depth estimation methods. Adding MAL into previous state-of-the-art methods leads to a reduction in depth estimation errors by up to 4.2\% and 10.8\% on KITTI and CityScapes benchmarks, respectively.
\end{abstract}

\section{introduction}
Accurate depth information is crucial for autonomous vehicles and robots to perceive and interact with environments in a manner akin to human cognition. Recent strides in deep learning methodologies have yielded remarkable progress in training networks to autonomously infer depth directly from RGB images. Expanding on this progress, a surge of interest has emerged in leveraging extensive, unlabeled real-world data, driving the pursuit of self-supervised methodologies employing monocular videos as input \cite{Zhou_Brown_Snavely_Lowe_2017,watson2021temporal,bangunharcana2023dualrefine}. 

Early methods employ self-supervision by making the foundational assumption of a static scene and framing the depth estimation task as a cross-view consistency problem \cite{Zhou_Brown_Snavely_Lowe_2017}, where the difference between the current frame and the reprojected frame from its neighbor serves as an image reprojection loss function. Recent state-of-the-art techniques employ multiple frames as input \cite{bangunharcana2023dualrefine, feng2022disentangling, guizilini2022multi}, incorporating a reprojection and matching process at feature level across adjacent frames for a better scene geometry understanding. 

Despite the advancements mentioned above, challenges still exist in dynamic scenes due to the violation of the static scene assumption. Moving objects introduce errors in feature matching and image reprojection loss computation. Some methods~\cite{watson2021temporal,bangunharcana2023dualrefine,guizilini2022multi,feng2022disentangling,zhang2023dyna} use teacher-student distillation with a single-frame depth network as a teacher to alleviate errors in feature matching, but errors in loss remain. Other approaches employ optical flow \cite{yin2018geonet} or 3D motion fields \cite{li2021unsupervised,lee2021attentive,hui2022rm,zhang2023dyna} to model object motion, or rely on semantic segmentation to separate foreground and background objects \cite{feng2022disentangling,klingner2020self}. However, these techniques often introduce complex algorithms into the network's forward pass, posing integration challenges with existing self-supervised depth estimation approaches. Our research aims to tackle these enduring challenges, enhancing the effectiveness of self-supervised depth estimation in the presence of dynamic elements while minimizing additional integration and inference costs.


\begin{figure}[t!]
    \centering
    \includegraphics[width=\linewidth]{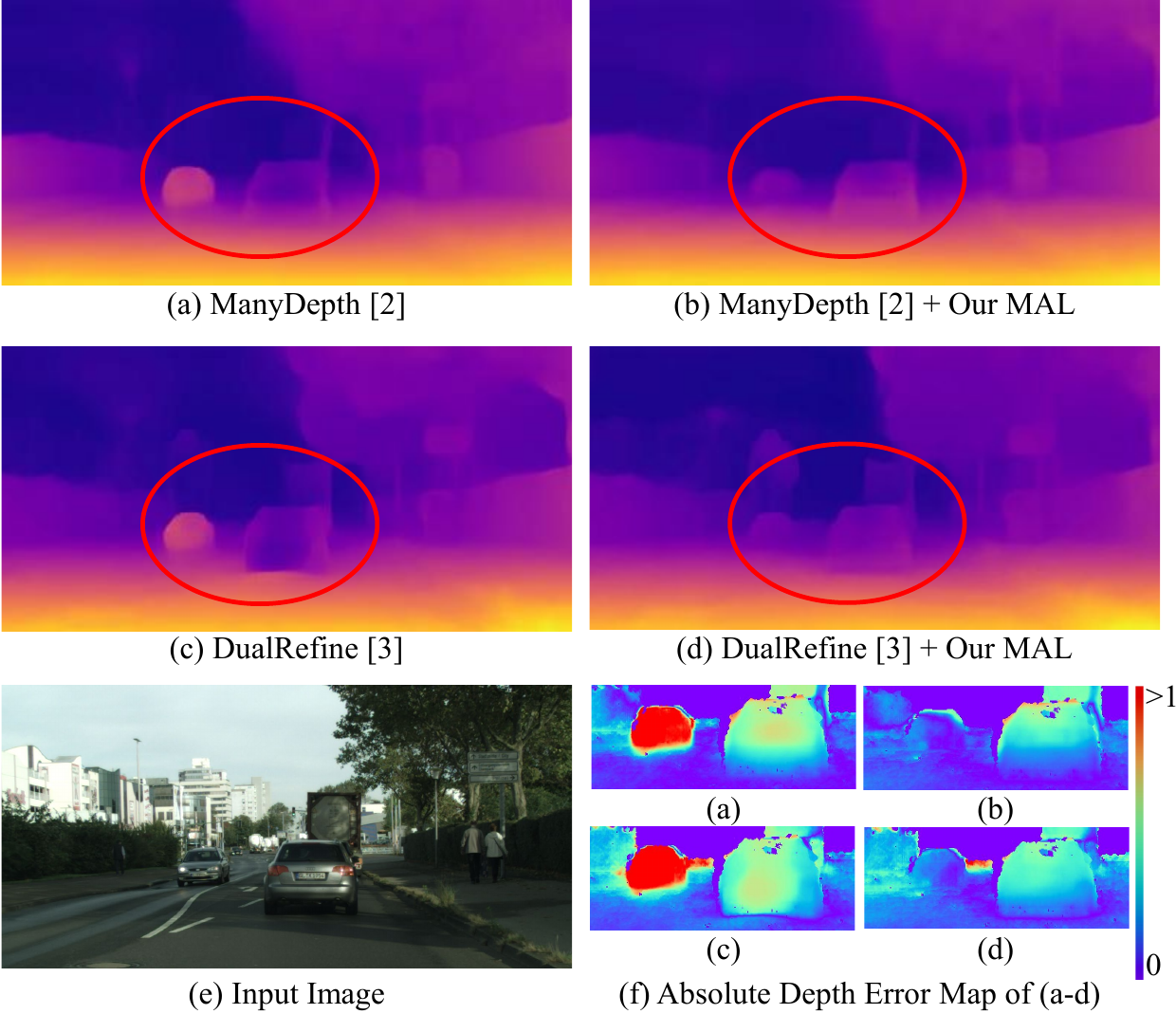}
    \caption{\textbf{Qualitative Demonstration of Our MAL's Effectiveness on CityScapes \cite{cordts2016cityscapes} Dataset}. MAL is designed for multi-frame depth estimation methods (a, c). It's a plug-and-play module (b, d) aimed at improving depth perception (f), especially for moving objects, in dynamic scenes (e).}
    \label{fig:poster}
    \vspace{-0.53cm}
\end{figure}

In this paper, we propose Motion-Aware Loss (MAL), a plug-and-play module designed for multi-frame self-supervised depth estimation from monocular videos. The primary aim is to enhance depth estimation in dynamic scenes through a novel approach to loss computation. We leverage temporal coherence in adjacent frames of monocular videos to address errors from moving objects in the image reprojection loss and enhance distillation in the teacher-student network to mitigate errors in the feature matching process. In a group of three consecutive frames, the antecedent and subsequent frames exhibit a symmetrical correspondence with regard to the central frame. Assuming uniform linear motion due to the short time interval between adjacent frames, we perform positional adjustments on dynamic elements and reconstruct occluded regions utilizing the symmetrical frame. This correction helps eliminate errors introduced by object motion in loss.
Meanwhile, previous methods \cite{watson2021temporal,bangunharcana2023dualrefine,feng2022disentangling,guizilini2022multi,zhang2023dyna} confine distillation operations to regions where the difference between the output of teacher network and the depth of the lowest feature matching cost in student network exceeds a prescribed threshold, and straightforwardly adopt the teacher's output as the distillation target. To further reduce errors caused by object motion in feature matching, we propose an extension of distillation across the entire image domain, advocating the utilization of the loss value as a criterion to select the more accurate depths between the outputs of the two networks as the distillation target.

This approach brings two key benefits. Firstly, this module is exclusively confined to the training phase, guaranteeing real-time inference efficiency. Secondly, as the modifications are restricted to the loss calculation stage, this module can be effortlessly and swiftly integrated into existing methods without the need for changes in the base model.

Our main contributions are:
\begin{itemize}
    \item We propose Motion-Aware Loss (MAL), a plug-and-play module to enhance multi-frame self-supervised depth estimation methods. It operates at the loss computation level, ensuring improved results without incurring additional computational overhead during inference.
    \item In MAL, we propose to leverage the temporal motion information inherent in neighboring frames and employ a new distillation scheme that spans the entire depth map. This strategic combination leads to notable enhancements in depth estimations, particularly in dynamic scenes.
    \item We integrated our MAL module into multiple multi-frame self-supervised depth estimation methods. Notably, we observed up to a remarkable 4.2\% improvement on KITTI and an impressive 10.8\% enhancement on CityScapes benchmarks, underscoring its efficacy.
\end{itemize}

\section{Related Work}
\subsection{Self-Supervised Depth Estimation}
Self-supervised depth estimation initially emerged as a technique for stereo pairs, where the estimated depth is constrained by a novel view synthesis process \cite{godard2017unsupervised}. In this context, two images of the same scene are captured from different positions, and one image can be synthesized with the other using the estimated depth based on Structure from Motion. This framework was later adapted to monocular settings, where monocular video sequences serve as input \cite{Zhou_Brown_Snavely_Lowe_2017}. A pose network is concurrently trained with the depth prediction network to model camera ego-motion. 
Previous advancements in this field include handling object occlusions \cite{monodepth2}, ensuring scale consistency across frames \cite{Mahjourian_Wicke_Angelova_2018,bian2019unsupervised, Chen_Schmid_Sminchisescu_2019,tiwari2020pseudo, wang2021can}, and improving network architectures \cite{guizilini20203d}.

\subsection{Self-Supervised Depth Estimation in Dynamic Scenes}
Monocular videos usually contain dynamic objects, which violate the static-scene assumption inherent to self-supervised depth estimation methodologies. To address this issue, some methods explicitly model pixel-wise motion using optical flow \cite{yin2018geonet} or 3D motion fields \cite{li2021unsupervised,lee2021attentive,hui2022rm}. Others leverage semantic cues \cite{lee2021attentive,casser2019unsupervised,lee2021learning}. They distinguish moving objects from the background and model object-level motion. Similarly, our MAL module also leverages instance segmentation information to address dynamic scenes.

\subsection{Multi-Frame Self-Supervised Depth Estimation}
Depth estimation from a single image is inherently challenging due to its ill-posed nature \cite{hartley2003multiple}. Consequently, recent research in self-supervised depth estimation has focused on multi-frame methods, which utilize multiple images during inference. ManyDepth \cite{watson2021temporal} introduces a feature matching scheme based on cost volume construction to leverage geometric information between frames. Building upon this approach, recent advancements integrate attention mechanisms into cost volume construction \cite{guizilini2022multi} and employ deep equilibrium models to improve the depth and pose estimates \cite{bangunharcana2023dualrefine}. DynamicDepth \cite{feng2022disentangling} leverages instance segmentation results to handle object motion by adjusting the positions of moving objects in input frames. However, DynamicDepth requires both the teacher network's estimated depth and the input frame modification during inference. Dyna-DepthFormer \cite{zhang2023dyna} utilizes self- and cross-attention modules to aggregate multi-frame and design a 3D motion field jointly trained with the depth network to handle moving objects. In contrast, our MAL module is solely involved in loss computation, ensuring that the real-time inference performance remains unaffected.
\section{Method}
\subsection{Framework of Self-Supervised Depth Estimation}

\begin{figure}[t!]
    \centering
    \includegraphics[width=\linewidth]{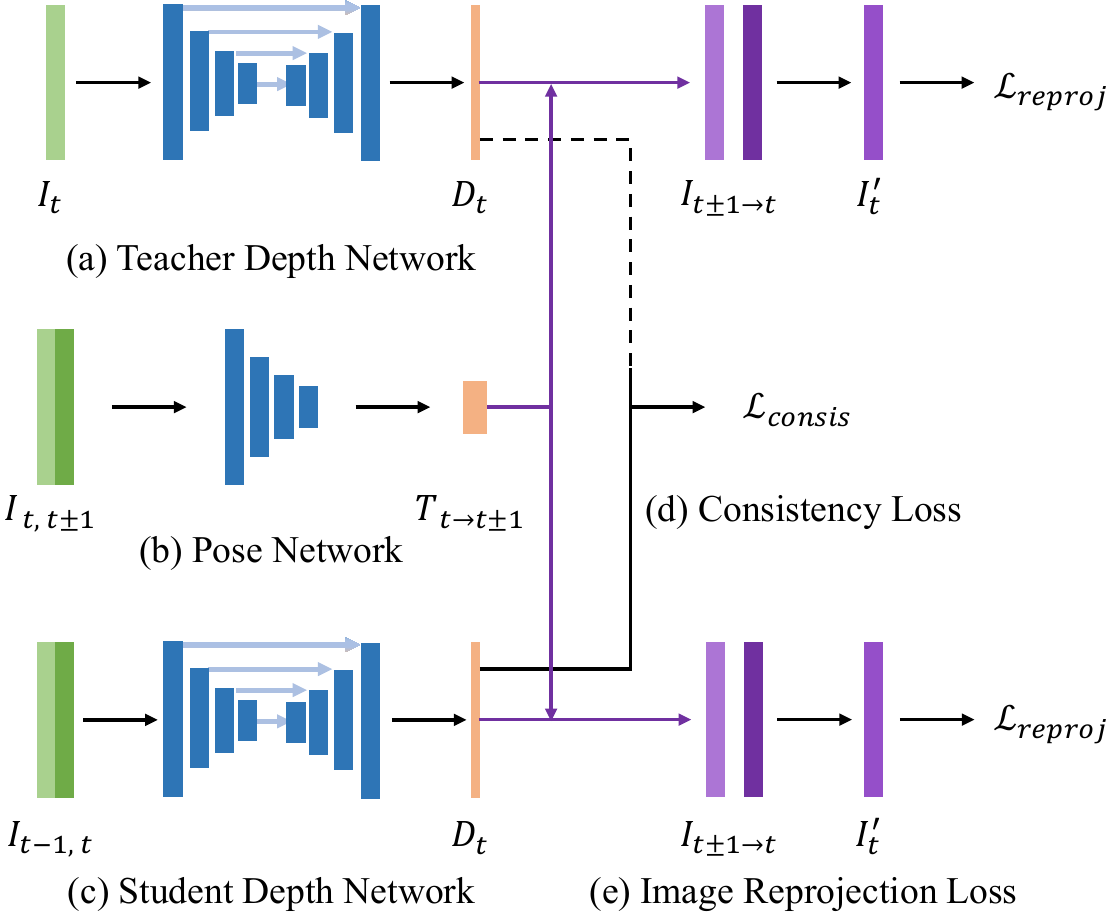}
    \caption{\textbf{Framework of Multi-Frame Self-Supervised Depth Estimation}. The three sub-networks (a-c) are trained concurrently with both image reprojection loss (e) and consistency loss (d). The dotted line indicates that the gradients of the teacher network are not updated by the consistency loss.}
    \label{fig:pipeline}
    \vspace{-0.5cm}
\end{figure}

Here, we revisit the multi-frame self-supervised depth estimation methodology \cite{watson2021temporal}. The framework (Fig. \ref{fig:pipeline}) consists of a teacher depth network, a student depth network, and a shared pose network. The teacher network generates a depth map from a single frame, while the student network uses two consecutive frames to predict the latter frame's depth map. Both networks share the same architecture, with a key distinction: the student constructs a cost volume to match features between adjacent frames in the encoder. This feature matching process is absent in the teacher network. The shared pose network estimates camera ego-motion as a six-dimensional vector, encompassing three dimensions for rotation angles and three for translation, and is used by both the teacher and student networks.

An image reprojection loss is used to train the framework (Fig. \ref{fig:pipeline} (e)). Denoting three consecutive frames as $I_{t-1}$, $I_{t}$, and $I_{t+1}$, we can project each pixel $p_t$ in $I_t$ onto $I_{t\pm 1}$ based on structure-from-motion under a static-scene assumption:
\begin{equation}\label{formula_warp}
    p_{t\pm 1}\sim KT_{t\to t\pm 1}D_{t}(p_{t})K^{-1}p_{t}
\end{equation}
where $p_{t\pm 1}$ are pixels in $I_{t\pm 1}$ and 
$K$ is camera intrinsic matrix. $D_{t}(p_{t})$ and $T_{t\to t\pm 1}$ are depth and camera pose predicted by the network. Pixels in $I_{t\pm 1}$ are sampled according to Eqn.~(\ref{formula_warp}) to reconstruct image at $I_t$'s viewpoint:
\begin{equation}
    I_{t\pm 1\to t}[p_{t}]=I_{t\pm 1}<p_{t\pm 1}>
\end{equation}
where $<>$ represents bilinear sampling.

The two reconstructed images, $I_{t\pm 1\to t}$, are combined pixel-wisely to handle occlusions. The final reconstructed image is created by choosing the pixel from either $I_{t-1\to t}$ or $I_{t+1\to t}$ with the lower photometric error compared to $I_t$ \cite{monodepth2}:
\begin{equation}
    \label{formula_baseline}
    I_{t}^{\prime} = \mathcal{P}(I_{t-1\to t}, I_{t+1\to t})
\end{equation}
where $\mathcal{P}$ represents the pixel selection and $I_{t}^{\prime}$ is the final reconstructed image. The photometric difference between $I_{t}^{\prime}$ and $I_{t}$ serves as the image reprojection loss. 

The feature matching process in the student network involves projecting features of $I_{t-1}$ to $I_t$'s viewpoint with pixel correspondences in Eqn. (\ref{formula_warp}). This projection employs a predefined set of uniformly distributed depth planes and seeks for depth to match the projected features with those of $I_t$. However, dynamic regions induce errors in this matching process, resulting in suboptimal depth estimates when training the student network solely with the image reprojection loss \cite{watson2021temporal}. To address this limitation, an asymmetric distillation scheme is employed, transferring knowledge from the teacher network, which does not involve the feature matching process, to the student. An uncertainty mask, denoted as $\mathcal{M}$, is computed through pixel-wise comparisons between the predicted depths of the teacher ($D_{t}$) and the depth with the lowest matching cost ($D_{cv}$) \cite{watson2021temporal}:
\begin{equation}\label{formula:mask}
\mathcal{M}=\max(\frac{D_{cv}-D_{t}}{D_{t}}, \frac{D_{t}-D_{cv}}{D_{cv}}) > 1
\end{equation}
During training, the reliable area ($\neg{\mathcal{M}}$) is supervised by the image reprojection loss, while the unreliable area ($\mathcal{M}$) is instead supervised by a consistency loss, calculated as the L1 difference between depths predicted by the teacher and student. Besides, we use an edge-aware smoothness loss $\mathcal{L}_s$ with a weight of $\lambda_{s}=1e-3$ as per standard practice \cite{monodepth2}:
\begin{equation}
    \mathcal{L}_{s} = |\partial_{x}d_{t}^{*}|e^{-|\theta_{x}I_{t}|} + |\partial_{y}d_{t}^{*}|e^{-|\theta_{y}I_{t}|}
\end{equation}
where $d_t^{*} = d_{t}/\overline{d_{t}}$ is the mean-normalized inverse depth. The original loss of the student network can be formulated as:
\begin{equation}\label{formula:lossori}
    \mathcal{L}_{ori} = \neg{\mathcal{M}} \cdot \mathcal{L}_{reproj} + \mathcal{M} \cdot \mathcal{L}_{consis} + \lambda_{s} \cdot \mathcal{L}_{s}
\end{equation}

\subsection{Temporal Hints}
\begin{figure}[t!]
    \centering
    \includegraphics[width=\linewidth]{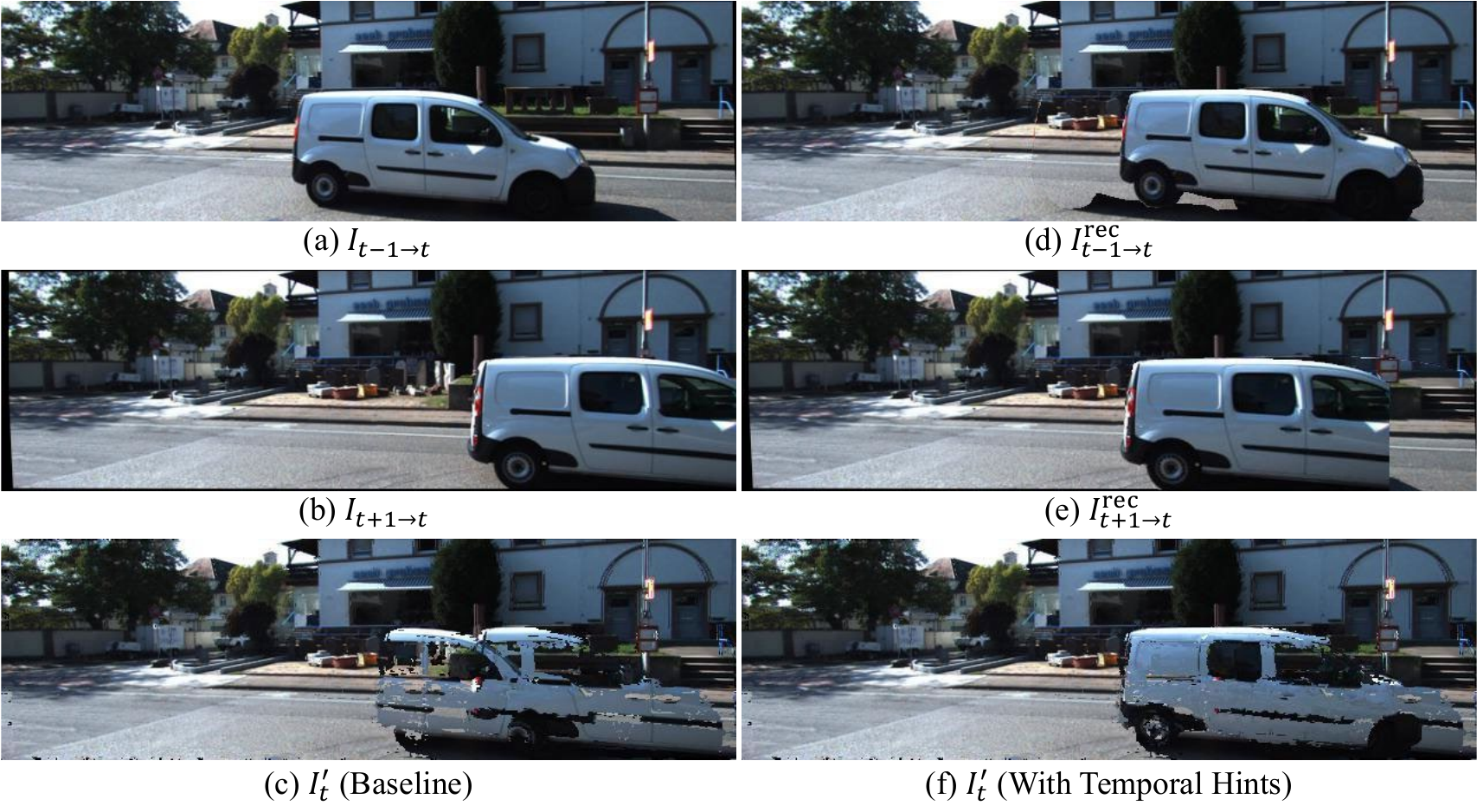}
    \caption{\textbf{Temporal Hints}. Linking object positions to input frames' temporal order via a linear motion model, we align object positions (d-e) and significantly reduce motion-induced errors in the reconstructed image (f).}
    \label{fig:displace}
    \vspace{-0.5cm}
\end{figure}
The reconstructed images $I_{t\pm 1\to t}$ represent images captured at times $t\pm1$ from the same viewpoint as $I_t$, as depicted in Fig. \ref{fig:displace} (a-b). In regions containing moving objects in $I_{t\pm 1\to t}$, photometric errors arise not from the estimated depth in $I_{t\pm 1\to t}$, but rather from inherent geometric changes. These errors subsequently affect the quality of the final reconstructed image, as demonstrated in Fig. \ref{fig:displace} (c). To tackle this challenge, we leverage temporal coherence in adjacent frames to adjust the positions of moving objects in $I_{t\pm 1\to t}$.

We initiate the process by employing a pre-trained instance segmentation model \cite{cheng2022masked} to identify moving objects, such as vehicles and pedestrians, within the scene. The parameters of this segmentation model don't update in training. To establish correspondences between instances across consecutive frames, we utilize the Hungarian algorithm. This algorithm leverages instance class labels and Intersection over Union (IoU) metrics between instance masks as the cost function.

When shooting monocular video, frames are captured with short time intervals. For instance, in the KITTI dataset \cite{geiger2012we}, data is recorded at 10Hz, resulting in a time gap of 0.1 seconds between frames. Consequently, we approximate the position of dynamic objects at time $t$ as the average of their positions in $I_{t\pm 1\to t}$. Considering the possibility of dynamic objects moving out of the camera's field of view, which may result in truncated objects near image borders in $I_{t\pm 1\to t}$, we implement a bounding-box-level object displacement calculation method to mitigate this issue.

We specifically concentrate on instances that are consistently present in both $t$$\pm$1 frames. This means that even if an object is partially truncated, either its left or right boundary should remain intact from $t$$-$1 to $t$$+$1, as depicted in Fig. \ref{fig:displace} (a-b). Hence, we approximate the horizontal displacement of the object as the maximum value between the displacement of its left boundary and right boundary from $t$$-$1 to $t$$+$1:
\begin{equation}\label{formula:displacemenet}
    \begin{aligned}
        \Delta h_{t-1\to t+1}^{i} &= \max(|l_{t+1}^{i}-{l}_{t-1}^{i}|, |{r}_{t+1}^{i}-{r}_{t-1}^{i}|) \\
    \end{aligned}
\end{equation}
Here ${l}_{t}^{i}$ and ${r}_{t}^{i}$ denote the left and right boundaries of instance $i$ at time $t$. The vertical displacement is calculated in a similar manner using the top and bottom boundaries of the instance. 

We align object positions in $I_{t\pm 1\to t}$ with those at time $t$ using the calculated displacement. This translation may uncover previously occluded areas. We leverage the symmetry between $t$$\pm$1 for restoration. Specifically, an area obscured by a moving object at time $t$$+$1 but exposed at time $t$ cannot be covered by the same object at time $t$$-$1. Therefore, we use pixels from $I_{t-1\to t}$ to fill in $I_{t+1\to t}^\mathrm{rec}$, and vice versa.

Due to the potential errors in the displacement calculations above, we introduce the motion-rectified image $I_{t\pm 1\to t}^\mathrm{rec}$ as an additional input in the final image reconstruction process:
\begin{equation}\label{formula:temp}
    I_{t}^{\prime} = \mathcal{P}(I_{t-1\to t}, I_{t+1\to t}, I_{t-1\to t}^\mathrm{rec}, I_{t+1\to t}^\mathrm{rec})
\end{equation}
Similar to Eqn.~(\ref{formula_baseline}), here, $\mathcal{P}$ represents the pixel selection operation and $I_{t}^{\prime}$ denotes the resulting reconstructed image. This approach effectively mitigates errors caused by object motion in both the reconstructed image and subsequent reprojection loss calculations. 

\subsection{Distillation Hints}
Besides the image reprojection loss computation, object motion also introduces errors in the feature matching progress inherent to the student network design. These errors cannot be easily mitigated by temporal hints alone because the feature matching occurs at the encoder, and the errors can propagate to subsequent parts of the student network.

To rectify these errors, we expand the distillation process within the region $\mathcal{M}$ mentioned in Eqn.~(\ref{formula:mask}) to cover the entire image, thereby maximizing the utilization of knowledge from the teacher network. We fuse the depth predictions from both the teacher and student networks on a pixel-wise basis, selecting the depth with a lower image reprojection loss to generate the target distillation depth map $D_{td}$. The distillation loss is computed as:
\begin{equation}\label{formula:lossdistil}
    \mathcal{L}_{distil} = \neg{\mathcal{M}} \cdot ||D_{s}-D_{td}||_{1}
\end{equation}
where $D_{s}$ represents the depth predicted by the student depth network. Similar to the settings in ManyDepth \cite{watson2021temporal}, the distillation process is unidirectional, and the teacher network does not update during the backward propagation of $\mathcal{L}_{distil}$.

\subsection{Loss Balancing}
The training of the student network is constrained with two loss terms in total: $\mathcal{L}_{ori}$, as computed by Eqn.~(\ref{formula:lossori}), and $ \mathcal{L}_{distil}$, computed by Eqn.~(\ref{formula:lossdistil}):
\begin{equation}\label{formula:lossours}
    \mathcal{L} = w_{1} \cdot \mathcal{L}_{ori} + w_{2} \cdot \mathcal{L}_{distil}
\end{equation}
To maintain an effective balance between the loss terms, we apply the multi-loss rebalancing algorithm (MLRA) \cite{lee2020multi}. Initially, each loss weight is set to 1/2, and these weights are iteratively updated during training based on the descending rate of each respective loss term. The hyperparameter $\lambda$ dictates whether the algorithm prioritizes rapidly descending loss terms or slower ones. 
\section{Experiments}
\begin{table*}[htbp]
    \centering
    \caption{\textbf{Depth Estimation Results on KITTI Eigen Split \cite{eigen2015predicting}.}}
    \scalebox{1.0}{
    \begin{tabular}{ccccccccccccc}
    \toprule
    \multicolumn{2}{c}{\multirow{2}{*}{Method}} & \multicolumn{1}{c}{\multirow{2}{*}{Test}} & \multicolumn{1}{c}{\multirow{2}{*}{Semantic}} & \multicolumn{1}{c}{\multirow{2}{*}{W$\times$H}} &
    \multicolumn{4}{c}{\emph{Errors}$\downarrow$} & \multicolumn {3}{c}{\emph{Accuracy}$\uparrow$}\\
    \cmidrule(r){6-9} \cmidrule(r){10-12}
    \multicolumn{2}{c}{} &  Frames & {} & & AbsRel & SqRel & RMSE & RMSE$_\mathrm{log}$ & $\delta<1.25$ & $\delta<1.25^{2}$ & $\delta<1.25^{3}$\\
    \midrule
    \multicolumn{2}{l}{Struct2Depth\cite{casser2019unsupervised}} & 1 &\CIRCLE &416$\times$128& 0.141 & 1.026 & 5.291 & 0.215 & 0.816 & 0.945 & 0.979\\ 
    \multicolumn{2}{l}{Bian \emph{et al.}\cite{bian2019unsupervised}} & 1 & &416$\times$128&  0.137 & 1.089 & 5.439 & 0.217 &  0.830 & 0.942 & 0.975\\
    \multicolumn{2}{l}{Gordon \emph{et al.}\cite{gordon2019depth}}  & 1 &\CIRCLE & 416$\times$128 & 0.128 & 0.959 & 5.230 & 0.212 & 0.845 & 0.947 & 0.976\\ 
    \multicolumn{2}{l}{MonoDepth2 \cite{monodepth2}}  & 1 & & 640$\times$192 & 0.115 & 0.903 & 4.863 & 0.193 & 0.877 & 0.959 & 0.981\\
    \multicolumn{2}{l}{InstaDM \cite{lee2021learning}}  &1 & \CIRCLE& 832$\times$256  & 0.112 & 0.777 & 4.772 & 0.191 & 0.872 & 0.959 & 0.982\\ 
    \multicolumn{2}{l}{Packnet-SFM \cite{guizilini20203d}}  & 1 & &640$\times$192  & 0.111 & 0.785 & 4.601 & 0.189 & 0.878 & 0.960 & 0.982\\ 
    \multicolumn{2}{l}{Wang \emph{et al.}\cite{wang2021can}}   & 1 & & 640$\times$192 &  0.109 & 0.779 & 4.641 & 0.186 & 0.883 & 0.962 & 0.982 \\
    \multicolumn{2}{l}{RM-Depth \cite{hui2022rm}}  & 1 & & 640$\times$192 & 0.108 & 0.710 & 4.513 & 0.183 & 0.884 & 0.964 & 0.983 \\ 
    
    \multicolumn{2}{l}{Johnston \emph{et al.} \cite{johnston2020self}}   & 1& & 640$\times$192 &  0.106 & 0.861 & 4.699 & 0.185 & 0.889 & 0.962 & 0.982\\
    
    \multicolumn{2}{l}{Guizilini \emph{et al.}\cite{GuiziliniHLAG20}}   &1 &\CIRCLE & 640$\times$192 &  0.102 & {0.698} & {4.381} & 0.178 & 0.896 & 0.964 & {0.984}\\
    
    \multicolumn{2}{l}{Wang \cite{wang2020self}}  & 2(-1,0) & & 640$\times$192 &  0.106 & 0.799 & 4.662 & 0.187 & {0.889} & {0.961} & {0.982}\\ 
    
    \multicolumn{2}{l}{DynamicDepth \cite{feng2022disentangling}} & 2(-1,0) & \CIRCLE & 640$\times$192 &  {0.096} & 0.720 & 4.458 & {0.175} & 0.897 &  0.964 & {0.984} \\
    \multicolumn{2}{l}{Dyna-DepthFormer \cite{zhang2023dyna}} & 2(-1,0)&  & 640$\times$192 &  {0.094} & 0.734 & 4.442 & {0.169} & 0.893 &  0.967 & {0.983} \\
    \multicolumn{2}{l}{DepthFormer \cite{guizilini2022multi}} & 2(-1,0)&  & 640$\times$192 & {0.090} & {0.661} & {4.149} & 0.175 & 0.905 & 0.967 & 0.984 \\

    \midrule
    \multicolumn{2}{l}{ManyDepth \cite{watson2021temporal}}  & 2(-1,0) & & 640$\times$192 &  0.098 & 0.770 & 4.459 & 0.176 & {0.900} & {0.965} & {0.983}\\ 
    \multicolumn{2}{l}{+MAL}  & 2(-1,0) & \CIRCLE & 640$\times$192 &  \textbf{0.094} & \textbf{0.732} & \textbf{4.425} & \textbf{0.174} & \textbf{0.906} & \textbf{0.966} & \textbf{0.983}\\ 
    \midrule
    \multicolumn{2}{l}{DualRefine \cite{bangunharcana2023dualrefine}} & 2(-1,0) & & 640$\times$192 & {0.087} & 0.698 & 4.234 & 0.170 & 0.914 & 0.967 & 0.983 \\
    \multicolumn{2}{l}{+MAL} & 2(-1,0) &\CIRCLE & 640$\times$192 & \textbf{0.087} & \textbf{0.690} & \textbf{4.227} & \textbf{0.169} & \textbf{0.915} & \textbf{0.968} & \textbf{0.983} \\ 
    \bottomrule
    \end{tabular}}
\label{tab:sota_k}
\end{table*}

\begin{table*}[ht]
    \centering
    \caption{\textbf{Depth Estimation Results on CityScapes} \cite{cordts2016cityscapes}.} 
    \scalebox{1.0}{
    \begin{tabular}{ccccccccccccc}
    \toprule
    \multicolumn{2}{c}{\multirow{2}{*}{Method}} & \multicolumn{1}{c}{\multirow{2}{*}{Test}} & \multicolumn{1}{c}{\multirow{2}{*}{Semantic}} & \multicolumn{1}{c}{\multirow{2}{*}{W$\times$H}} &
    \multicolumn{4}{c}{\emph{Errors}$\downarrow$} & \multicolumn {3}{c}{\emph{Accuracy}$\uparrow$}\\
    \cmidrule(r){6-9} \cmidrule(r){10-12}
    \multicolumn{2}{c}{} & Frames & & & AbsRel & SqRel & RMSE & RMSE$_\mathrm{log}$ & $\delta<1.25$ & $\delta<1.25^{2}$ & $\delta<1.25^{3}$\\
    \midrule
    \multicolumn{2}{l}{Struct2Depth \cite{casser2019unsupervised}} & 1 & \CIRCLE & 416$\times$128 & 0.145 & 1.737 & 7.280 & 0.205 & 0.813 & 0.942 & 0.976\\
    \multicolumn{2}{l}{MonoDepth2 \cite{monodepth2}} & 1& & 416$\times$128 & 0.129 & 1.569 & 6.876 & 0.187 & 0.849 & 0.957 & 0.983\\
    \multicolumn{2}{l}{Gordon \emph{et al.} \cite{gordon2019depth}} &1 & \CIRCLE& 416$\times$128 & 0.127 & 1.330 & 6.960 & 0.195 & 0.830 & 0.947 & 0.981\\
    \multicolumn{2}{l}{Li \emph{et al.} \cite{li2021unsupervised}} &1 &  & 416$\times$128 & 0.119 & 1.290 & 6.980 & 0.190 & 0.846 & 0.952 & 0.982\\
    \multicolumn{2}{l}{InstaDM \cite{lee2021learning}} & 1 & \CIRCLE & 832$\times$256 & 0.111 & 1.158 & 6.437 & 0.182 & 0.868 & 0.961 & 0.983\\
    \multicolumn{2}{l}{RM-Depth \cite{hui2022rm}} & 1 & & 416$\times$128 & 0.100 & {0.839} & 5.774 & 0.154 & 0.895 & 0.976 & {0.993} \\
    \multicolumn{2}{l}{Dyna-DepthFormer \cite{zhang2023dyna}} &  2(-1,0) & & 416$\times$128 &  {0.100} & 0.834 & 5.843 & {0.154} & 0.901 &  0.975 & {0.992} \\
    
    \midrule    
    \multicolumn{2}{l}{ManyDepth \cite{watson2021temporal}} & 2(-1, 0) & & 416$\times$128 & 0.114 & 1.193 & 6.223 & 0.170 & 0.875 & 0.967 & 0.989\\
    \multicolumn{2}{l}{+MAL} & 2(-1, 0) & \CIRCLE & 416$\times$128 & \textbf{0.103} & \textbf{1.073} & \textbf{5.952} & \textbf{0.157} & \textbf{0.896} & \textbf{0.973} & \textbf{0.991} \\ 
    \midrule
    \multicolumn{2}{l}{DynamicDepth \cite{feng2022disentangling} (paper)} & 2(-1, 0) & \CIRCLE & 416$\times$128 & 0.103 & 1.000 & 5.867 & 0.157 & 0.895 &  0.974 & 0.991 \\
    \multicolumn{2}{l}{Officially Provided Model} & 2(-1, 0) & \CIRCLE & 416$\times$128 & 0.104 & 1.011 & 5.987 & 0.159 & 0.890 &  0.972 & 0.991 \\
    \multicolumn{2}{l}{+MAL} & 2(-1, 0) &\CIRCLE & 416$\times$128 & \textbf{0.101} & \textbf{0.957} & \textbf{5.865} & \textbf{0.156} & \textbf{0.895} & \textbf{0.974} & \textbf{0.991}\\ 
    \midrule
    
    \multicolumn{2}{l}{DualRefine \cite{bangunharcana2023dualrefine}} & 2(-1, 0) & & 416$\times$128 & 0.111 & 1.248 & 6.035 & 0.164 & 0.896 & 0.971 & 0.989 \\ 
    \multicolumn{2}{l}{+MAL} & 2(-1, 0) &\CIRCLE & 416$\times$128 & \textbf{0.099} & \textbf{0.973} & \textbf{5.530} & \textbf{0.149} & \textbf{0.905} & \textbf{0.977} & \textbf{0.992}\\ 
    \bottomrule
\end{tabular}}
\label{tab:sota_cs}
\end{table*}
\begin{table*}[ht]
    \centering
    \caption{\textbf{Depth Estimation Results on CityScapes} \cite{cordts2016cityscapes}.} 
    \scalebox{1.0}{
    \begin{tabular}{ccccccccccccc}
    \toprule
    \multicolumn{2}{c}{\multirow{2}{*}{Method}} & \multicolumn{1}{c}{\multirow{2}{*}{Test}} & \multicolumn{1}{c}{\multirow{2}{*}{Semantic}} & \multicolumn{1}{c}{\multirow{2}{*}{W$\times$H}} &
    \multicolumn{4}{c}{\emph{Errors}$\downarrow$} & \multicolumn {3}{c}{\emph{Accuracy}$\uparrow$}\\
    \cmidrule(r){6-9} \cmidrule(r){10-12}
    \multicolumn{2}{c}{} & Frames & & & AbsRel & SqRel & RMSE & RMSE$_\mathrm{log}$ & $\delta<1.25$ & $\delta<1.25^{2}$ & $\delta<1.25^{3}$\\
    \midrule
    \multicolumn{2}{l}{Struct2Depth \cite{casser2019unsupervised}} & 1 & \CIRCLE & 416$\times$128 & 0.145 & 1.737 & 7.280 & 0.205 & 0.813 & 0.942 & 0.976\\
    \multicolumn{2}{l}{MonoDepth2 \cite{monodepth2}} & 1& & 416$\times$128 & 0.129 & 1.569 & 6.876 & 0.187 & 0.849 & 0.957 & 0.983\\
    \multicolumn{2}{l}{Gordon \emph{et al.} \cite{gordon2019depth}} &1 & \CIRCLE& 416$\times$128 & 0.127 & 1.330 & 6.960 & 0.195 & 0.830 & 0.947 & 0.981\\
    \multicolumn{2}{l}{Li \emph{et al.} \cite{li2021unsupervised}} &1 &  & 416$\times$128 & 0.119 & 1.290 & 6.980 & 0.190 & 0.846 & 0.952 & 0.982\\
    \multicolumn{2}{l}{InstaDM \cite{lee2021learning}} & 1 & \CIRCLE & 832$\times$256 & 0.111 & 1.158 & 6.437 & 0.182 & 0.868 & 0.961 & 0.983\\
    \multicolumn{2}{l}{RM-Depth \cite{hui2022rm}} & 1 & & 416$\times$128 & 0.100 & {0.839} & 5.774 & 0.154 & 0.895 & 0.976 & {0.993} \\
    \multicolumn{2}{l}{Dyna-DepthFormer \cite{zhang2023dyna}} &  2(-1,0) & & 416$\times$128 &  {0.100} & 0.834 & 5.843 & {0.154} & 0.901 &  0.975 & {0.992} \\
    
    \midrule    
    \multicolumn{2}{l}{ManyDepth \cite{watson2021temporal}} & 2(-1, 0) & & 416$\times$128 & 0.114 & 1.193 & 6.223 & 0.170 & 0.875 & 0.967 & 0.989\\
    \multicolumn{2}{l}{+MAL} & 2(-1, 0) & \CIRCLE & 416$\times$128 & \textbf{0.103} & \textbf{1.073} & \textbf{5.952} & \textbf{0.157} & \textbf{0.896} & \textbf{0.973} & \textbf{0.991} \\ 
    \midrule
    \multicolumn{2}{l}{DynamicDepth \cite{feng2022disentangling} (paper)} & 2(-1, 0) & \CIRCLE & 416$\times$128 & 0.103 & 1.000 & 5.867 & 0.157 & 0.895 &  0.974 & 0.991 \\
    \multicolumn{2}{l}{Officially Provided Model} & 2(-1, 0) & \CIRCLE & 416$\times$128 & 0.104 & 1.011 & 5.987 & 0.159 & 0.890 &  0.972 & 0.991 \\
    \multicolumn{2}{l}{+MAL} & 2(-1, 0) &\CIRCLE & 416$\times$128 & \textbf{0.101} & \textbf{0.957} & \textbf{5.865} & \textbf{0.156} & \textbf{0.895} & \textbf{0.974} & \textbf{0.991}\\ 
    \midrule
    
    \multicolumn{2}{l}{DualRefine \cite{bangunharcana2023dualrefine}} & 2(-1, 0) & & 416$\times$128 & 0.111 & 1.248 & 6.035 & 0.164 & 0.896 & 0.971 & 0.989 \\ 
    \multicolumn{2}{l}{+MAL} & 2(-1, 0) &\CIRCLE & 416$\times$128 & \textbf{0.099} & \textbf{0.973} & \textbf{5.530} & \textbf{0.149} & \textbf{0.905} & \textbf{0.977} & \textbf{0.992}\\ 
    \bottomrule
\end{tabular}}
\label{tab:sota_cs}
\end{table*}

\begin{table*}[ht]
    \centering
    \caption{\textbf{Ablation Study for ManyDepth on CityScapes \cite{cordts2016cityscapes}}.}
    \scalebox{1.0}{
    \begin{tabular}{ccccccccc}
        \toprule
        {\multirow{2}{*}{Method}} & {\multirow{2}{*}{Loss Terms}} & \multicolumn{4}{c}{\emph{Errors}$\downarrow$} & \multicolumn {3}{c}{\emph{Accuracy}$\uparrow$}\\
        \cmidrule(r){3-6} \cmidrule(r){7-9} 
        & Combination & AbsRel & SqRel & RMSE & RMSE$_\mathrm{log}$ & $\delta<1.25$ & $\delta<1.25^{2}$ & $\delta<1.25^{3}$\\
        \midrule
        ManyDepth~\cite{watson2021temporal} & Original & 0.114 & 1.193 & 6.223 & 0.170 & 0.875 & 0.967 & 0.989 \\
        \midrule
        +Temporal Hints & Original & 0.111 & 1.182 & 6.127 & 0.165 & 0.882 & 0.970 & 0.990 \\ 
        +Distillation Hints & Sum Up & 0.114 & 1.300 & 6.296& 0.168 & 0.882 & 0.969 & 0.989 \\ 
        +Distillation Hints & MLRA \cite{lee2020multi} & 0.111 & 1.179 & 6.083 & 0.164 & 0.883 & 0.970 & 0.990 \\ 
        \midrule
        +MAL & Sum Up & 0.109 &  1.141 & 6.035 & 0.162 & 0.887 & 0.971 & 0.990 \\ 
        +MAL & MLRA & \textbf{0.103} & \textbf{1.073} & \textbf{5.952} & \textbf{0.157} & \textbf{0.896} & \textbf{0.973} & \textbf{0.991} \\ 
        
        \bottomrule    
    \end{tabular}}
\label{tab:ablation}
\end{table*}

\subsection{Dataset}
\subsubsection{KITTI}
KITTI \cite{geiger2012we} is the standard benchmark for self-supervised depth estimation evaluation. It is an autonomous driving dataset featuring urban scenes. Following the established practice in previous work, we use the data split of Eigen \cite{eigen2014depth} and the data pre-processing to remove static frames established by \cite{Zhou_Brown_Snavely_Lowe_2017}, resulting in 39,810 monocular triplets for training, 4,424 for validation, and 697 for testing. 
\subsubsection{CityScapes}
CityScapes \cite{cordts2016cityscapes} is also a popular benchmark including numerous dynamic scenes with multiple moving objects \cite{casser2019unsupervised}. It is a notable benchmark for algorithms dealing with dynamic objects \cite{feng2022disentangling,li2021unsupervised,hui2022rm,lee2021learning}. We follow the protocol in previous work \cite{watson2021temporal, feng2022disentangling} and evaluate 1,525 images.

\subsection{Experiment Setup}
Currently, there are five prominent multi-frame self-supervised depth estimation methods in the literature \cite{watson2021temporal, guizilini2022multi, feng2022disentangling, bangunharcana2023dualrefine, zhang2023dyna}. These methods all incorporate image reprojection loss and the teacher-student distillation scheme in their architectures, which theoretically aligns them with our MAL module. However, as Dyna-DepthFormer\cite{zhang2023dyna} is not open-sourced and the training configuration file for DepthFormer\cite{guizilini2022multi} is not yet publicly accessible, we have chosen to evaluate MAL using ManyDepth \cite{watson2021temporal}, DynamicDepth \cite{feng2022disentangling}, and DualRefine \cite{bangunharcana2023dualrefine} as our baseline frameworks. Notably, DualRefine currently stands as the top-performing model on the KITTI benchmark, while DynamicDepth leads among multi-frame methods on the CityScapes benchmark, with the exception of the most recent Dyna-DepthFormer.

Since self-supervised learning predicts relative depth, we adhere to the single-image median scaling and cap depth values at 80 meters during evaluations, as is standard in the field \cite{monodepth2}. We assess the depth predictions using established depth evaluation metrics \cite{eigen2015predicting}, including Absolute Relative Error (AbsRel), Squared Relative Error (SqRel), Root Mean Squared Error (RMSE), Root Mean Squared Log Error (RMSElog), and accuracy within specified thresholds ($\delta$).

For ManyDepth+MAL and DynamicDepth+MAL, we fine-tune the official models provided by the authors from their respective GitHub repositories, employing a batch size of 24 and 12 with a learning rate of 1e-4 and 1e-5 respectively. DualRefine+MAL undergoes fine-tuning on KITTI with a batch size of 16 and a learning rate of 1e-5. In the case of CityScapes, where no pre-trained DualRefine model is available, we train DualRefine+MAL from scratch, utilizing a batch size of 16 and a learning rate of 1e-4 for 10 epochs. We employ the Adam optimizer \cite{kingma2014adam} with $\beta_{1}=0.9$ and $\beta_{2}=0.999$ across all experiments. $\lambda$ of MLRA is set to 3 and linearly decreases to -3.

\begin{figure*}[t!]
    \centering
    \includegraphics[width=\textwidth]{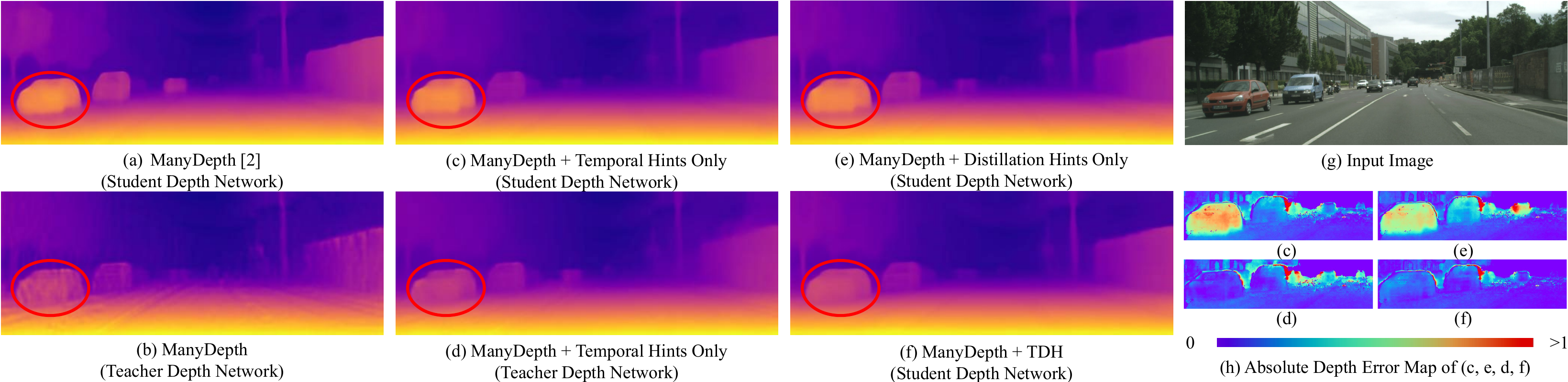}
    \caption{\textbf{Qualitative Analysis of the Indispensability of Both the Temporal and Distillation Hints}. Please refer to Section \ref{sec:complementary} for a detailed analysis.}
    \label{fig:distil}
    \vspace{-0.3cm}
\end{figure*}

\subsection{Evaluation Results}
We evaluate our method on KITTI and CityScapes benchmarks and the results are shown in Table \ref{tab:sota_k} and Table \ref{tab:sota_cs}. The \emph{Test Frames} column indicates the number of input frames during inference. A value of 1 corresponds to a single-frame method, while 2 (-1, 0) signifies a multi-frame method that employs the current frame and its previous frame as input.

According to statistics from previous work \cite{feng2022disentangling}, dynamic category objects (such as vehicles, pedestrians, cyclists) account for only 0.34\% of the pixels in the KITTI dataset, and most of the vehicles are stationary. Hence, previous state-of-the-art methods that specifically target dynamic objects \cite{feng2022disentangling,zhang2023dyna,hui2022rm} show a relative minor advantage on KITTI compared to CityScapes, where dynamic scenes are more prevalent. As for MAL, ManyDepth shows an improvement of up to 4.2\%, while adding MAL to DualRefine leads to higher $\delta <1.25$ and $\delta <1.25^{2}$, indicating a larger proportion of accurate inliers.

Meanwhile, on CityScapes, our MAL consistently enhances all seven depth evaluation metrics of ManyDepth and DualRefine as shown in Table \ref{tab:sota_cs}. Specifically, ManyDepth's depth estimation results can be improved by up to 10.1\%, and DualRefine demonstrates an enhancement of up to 10.8\%.

Further, we assess the impact of MAL on existing dynamic-scene-oriented methods, which also employ image reprojection loss and the teacher-student framework and is thus applicable for MAL. Since the code of Dyna-DepthFormer \cite{zhang2023dyna} and pre-computed masks of DynamicDepth \cite{feng2022disentangling} for KITTI are not publicly available, we experiment with DynamicDepth+MAL on CityScapes. Despite DynamicDepth's pre-optimized architecture for dynamic scenes, compared to the model officially provided by the authors of DynamicDepth, applying MAL yields a noticeable 5.34\% decrease in SqRel and an increase from 89.0\% to 89.5\% in the accuracy metric $\delta <1.25$, indicating a higher percentage of accurate inliers and a reduced proportion of outliers.

Our MAL offers a substantial performance improvement for existing multi-frame methods, achieving results comparable to state-of-the-art approaches. Importantly, MAL optimizes the algorithm at the loss level, making it easy to integrate into these established methods. Compared to other methods like RM-Depth \cite{hui2022rm}, DynamicDepth \cite{feng2022disentangling}, and Dyna-DepthFormer \cite{zhang2023dyna}, which are designed with specific network forward pass algorithms to address dynamic objects, MAL exhibits greater portability.

\subsection{Ablation Study}
We conduct an ablation study on CityScapes to dissect the contributions of each component of our MAL (Table \ref{tab:ablation}). Our baseline is ManyDepth \cite{watson2021temporal}. It is worth noting that our MAL enhances depth perception, even in the absence of MLRA (as demonstrated in the fourth row of Table \ref{tab:ablation}). MLRA plays a role in automatically generating more sensible weights, thereby providing an additional boost in performance.
\subsection{Qualitative Analysis}\label{sec:complementary}
Here we elucidate the indispensable roles of each component in our MAL. 
We disable distillation hints, only enable temporal hints, and employ the original loss function in Eqn. (\ref{formula:lossori}). In this case, it is noteworthy that moving objects can induce errors in the cost-volume-based feature matching process in the encoder of the student depth network, which may propagate and degrade its final output despite the temporal hints. Fig. \ref{fig:distil} (c, h) manifest an obvious error in the student's depth prediction for the car marked by the red circle. Conversely, the teacher network provides a notably more accurate prediction for this car (Fig. \ref{fig:distil} (d, h)), outperforming the baseline (Fig. \ref{fig:distil} (b)). This highlights the effectiveness of our temporal hints in enhancing depth perception for dynamic objects in cases where the feature matching process is absent. Moreover, it implies the traditional distillation scheme may not fully exploit the information in the teacher network to improve the student's performance.

Further, we enable both the temporal and the distillation hints. The depth prediction of this car becomes much better (Fig. \ref{fig:distil} (f, h)). Meanwhile, we disable the temporal hints and enable the distillation hints only (Fig. \ref{fig:distil} (e)). Even in the absence of temporal hints, the depth estimation for the circled car is superior to the case only with temporal hints (Fig. \ref{fig:distil} (h)(c) and (h)(e)). This underscore the efficacy of our distillation hints in facilitating a more effective information transfer from the teacher network to the student, compensating for the errors from the feature matching process in the student. However, with only distillation hints, the cars positioned behind the red-circled car show less accurate depth estimation (Fig. \ref{fig:distil} (h)(c), indicating that both the temporal and distillation hints are indispensable. 


\section{Conclusion}
In this paper, we present MAL, a plug-and-play module designed to augment depth perception, especially in dynamic scenes, using temporal and distillation hints. MAL can be seamlessly integrated with multi-frame self-supervised depth estimation methods and functions at the loss computation level, ensuring no additional inference time overhead. Our experimental results demonstrate that incorporating MAL into established multi-frame methods yields substantial improvements in depth estimation performance across the KITTI and CityScapes benchmarks.


\bibliographystyle{IEEEtran}
\bibliography{icra24}
\end{document}